\pdfoutput=1
\documentclass[journal]{IEEEtran}

\usepackage{amsmath,amsthm,amsfonts}
\usepackage{array}
\usepackage{algorithm}
\usepackage{algorithmic}
\usepackage{graphicx}
\usepackage{multirow}
\usepackage{balance}
\usepackage{url}

\newtheorem{definition}{Definition}

\newtheorem{lemma}{Lemma}
\newtheorem{proposition}{Proposition}
\newtheorem{example}{Example}

\ifCLASSINFOpdf
\else
\fi
\begin{document}
%
\title{Parallelized Tensor Train Learning of Polynomial Classifiers}
%
%
%

\author{Zhongming Chen$^{*}$, Kim~Batselier$^\dagger$, Johan A.K. Suykens$^\ddagger$ and Ngai~Wong$^\dagger$
\thanks{$^*$Department of Mathematics, School of Science, Hangzhou Dianzi University, Hangzhou 310018, China. Email: czm183015@126.com.}
\thanks{$^\dagger$Department of Electrical and Electronic Engineering, The University of Hong Kong. Email: \{kimb, nwong\}@eee.hku.hk.}
\thanks{$^\ddagger$KU Leuven, ESAT, STADIUS. B-3001 Leuven, Belgium. Email: johan.suykens@esat.kuleuven.be.}
}

\maketitle

\begin{abstract}
In pattern classification, polynomial classifiers are well-studied methods as they are capable of generating complex decision surfaces. Unfortunately, the use of multivariate polynomials is limited to kernels as in support vector machines, because polynomials quickly become impractical for high-dimensional problems. In this paper, we effectively overcome the curse of dimensionality by employing the tensor train format to represent a polynomial classifier. Based on the structure of tensor trains, two learning algorithms are proposed which involve solving different optimization problems of low computational complexity. Furthermore, we show how both regularization to prevent overfitting and parallelization, which enables the use of large training sets, are incorporated into these methods. The efficiency and efficacy of our tensor-based polynomial classifier are then demonstrated on the two popular datasets USPS and MNIST.



\end{abstract}


%
\IEEEpeerreviewmaketitle

\section{Introduction}
%
%
%

%

Pattern classification is the machine learning task of identifying to which category a new observation belongs, on the basis of a training set of observations whose category membership is known. 
This type of machine learning algorithm that uses a known training dataset to make predictions is called supervised learning, which has been extensively studied and has wide applications in the fields of bioinformatics \cite{SC06}, computer-aided diagnosis (CAD) \cite{CCCHL03}, machine vision \cite{R09}, speech recognition \cite{JHL97}, handwriting recognition \cite{XKS92}, spam detection and many others \cite{DHS12,zhang2014fruit,zhang2015}. Usually, different kinds of learning methods use different models to generalize from training examples to novel test examples. 

As pointed out in \cite{LBBH98,DS02}, one of the important invariants in these applications is the {\it local structure}: variables that are spatially or temporally nearby are highly correlated. Local correlations benefit extracting local features because configurations of neighboring variables can be classified into a small number of categories (e.g. edges, corners...).
For instance, in handwritten character recognition, correlations between image pixels that are nearby tend to be more reliable than the ones of distant pixels. Learning methods incorporating this kind of prior knowledge often demonstrate state-of-the-art performance in practical applications. One popular method for handwritten character recognition is using convolutional neural networks (CNNs) \cite{LBDHHHJ89, KSH12} which are variations of multilayer perceptrons designed to use minimal amounts of preprocessing. In this model, each unit in a layer receives inputs from a set of units located in a small neighborhood in the previous layer, and these mappings share the same weight vector and bias in a given convolutional layer. An important component of a CNN are the pooling layers, which implement a nonlinear form of down-sampling. In this way, the amount of parameters and computational load are reduced in the network.
Another popular method uses support vector machines (SVMs) \cite{CV95, CHCRL10}. The original finite-dimensional feature space is mapped into a much higher-dimensional space, where the inner product is easily computed through the `kernel trick'. By considering the Wolfe dual representation, one can find the maximum-margin hyperplane to separate the examples of different categories in that space. 
However, it is worth mentioning that these models require a large amount of memory and a long processing time to train the parameters. For instance, if there are thousands of nodes in the CNN, the weight matrices of fully-connected layers are of the order of millions. The major limitation of basic SVMs is the high computational complexity which is at least quadratic with the dataset size. One way to deal with large datasets in the SVM-framework is by using a fixed-size least squares SVM (fixed-size LS-SVM)~\cite{SVDDV02}, which approximates the kernel mapping in such a way that the problem can be solved in the primal space. 

\section{Tensors in machine learning}
Tensors are a multidimensional generalization of matrices to higher orders and have recently gained attention in the field of machine learning. The classification via tensors, as opposed to matrices or vectors, was first considered in \cite{SDDS12}, by extending the concept of spectral regularization for matrices to tensors. The tensor data is assumed to satisfy a particular low-rank Tucker decomposition, which unfortunately still suffers from an exponential storage complexity. Other work has focused speeding-up the convolution operation in CNNs~\cite{LGROL14} by approximating this operation with a low-rank polyadic decomposition of a tensor. In \cite{NPOV15}, the weight matrices of fully-connected layers of neural networks are represented by tensor trains (TTs), effectively reducing the number of parameters. TTs have also been used to represent nonlinear predictors~\cite{NTO16} and classifiers~\cite{SS16}. The key idea here is always to approximate a mapping that is determined by an exponential number of parameters $n^d$ by a TT with a storage complexity of $dnr^2$ parameters. To our knowledge, this idea has not yet been applied to polynomial classifiers that also suffer from the curse of dimensionality. The usual approach to circumvent the exponential number of polynomial coefficients would be to use SVMs with a polynomial kernel and solve the problem in the dual space. In this article, we exploit the efficient representation of a multivariate polynomial as a TT in order to avoid the curse of dimensionality, allowing us to work directly in the feature space. The main contributions are:

\begin{itemize}
\item We derive a compact description of a polynomial classifier using the TT format, avoiding the curse of dimensionality.

\item Two efficient learning algorithms are proposed by exploiting the TT structure.

\item Both regularization and a parallel implementation are incorporated into our methods, thus avoiding overfitting and allowing the use of large training datasets.

\end{itemize}

This paper is organized as follows. In Section~\ref{sec:pre}, we give a brief introduction to tensor basics, including the TT decomposition, important tensor operations and properties. The framework of TT learning for pattern classification is presented in Section~\ref{sec:TTLearning}. Based on different loss functions, two efficient learning algorithms are proposed in Section~\ref{sec:LearningAlg}, together with a discussion on regularization and parallelization. In Section~\ref{sec:experiments}, we test our algorithms on two popular datasets: USPS and MNIST and compare their performance with polynomial classifiers trained with LS-SVMs~\cite{SVDDV02}. Finally, some conclusions and further work are summarized in Section~\ref{sec:conclusion}.   

Throughout this paper, we use small letters $x, y, \ldots,$ for scalars, small bold letters ${\bf x}, {\bf y}, \ldots,$ for vectors, capital letters $A, B, \ldots,$ for matrices, and calligraphic letters ${\cal A}, {\cal B}, \ldots,$ for tensors. The transpose of a matrix $A$ or vector ${\bf x}$ is denoted by $A^\top$ and ${\bf x}^\top$, respectively. The identity matrix of dimension $n$ is denoted by $I_n$. A list of abbreviations used here is summarized in Table~\ref{List_abb}.
\begin{table}[!h]
\caption{List of Abbreviations}
\label{List_abb}
\centering
\begin{tabular}{rl}
\hline  \vspace{-2mm}   \\ 
TT    &  Tensor Train                                  \\
CNN   &  Convolutional Neural Network                  \\
SVM   &  Support Vector Machine                        \\
TTLS  &  Tensor Train learning by Least Squares         \\
TTLR  &  Tensor Train learning by Logistic Regression  \\
USPS  &  US Postal Service database                    \\
MNIST &  Modified NIST database                        \\
\hline
\end{tabular}
\end{table}

\section{Preliminaries}
\label{sec:pre}
\subsection{Tensors and pure-power-$\bf n$ polynomials}
\label{sec:sub:tensor}

\begin{figure}[!t]
\centering
\includegraphics[scale=0.9]{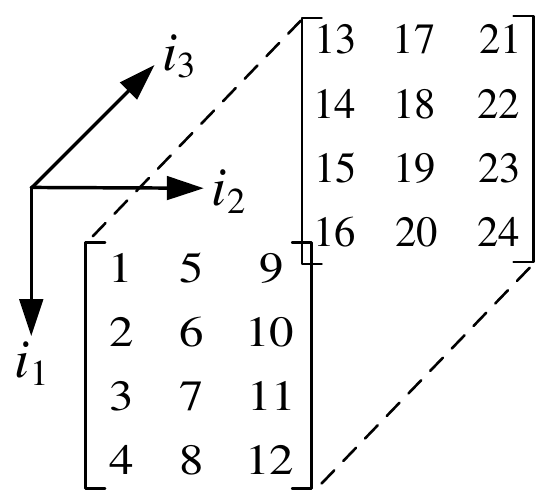}
\caption{An example tensor ${\cal A} =({\cal A}_{i_1 i_2 i_3}) \in \mathbb{R}^{4\times3\times2}$, where $i_1, i_2, i_3$ denote the indices for each mode respectively.}
\label{fig:tensor_example}
\end{figure}

A real $d$th-order or $d$-way tensor is a multidimensional array ${\cal A} \in \mathbb{R}^{n_1 \times n_2 \times \cdots \times n_d}$ that generalizes the notions of vectors and matrices to higher orders. Each of the entries $ {\cal A}_{i_1 i_2 \cdots i_d}$ is determined by $d$ indices. The numbers $n_1, n_2, \ldots, n_d$ are called the dimensions of the tensor. An example tensor with dimensions 4, 3, 2 is shown in Fig.~\ref{fig:tensor_example}. We now give a brief introduction to some required tensor operations and properties, more information can be found in \cite{KB09}.

The $k$-mode product ${\cal B} = {\cal A} \times_k U$ of a tensor ${\cal A} \in \mathbb{R}^{n_1 \times n_2 \times \cdots \times n_d}$ and a matrix $U \in \mathbb{R}^{n_k' \times n_k}$ is defined by
\begin{equation}\label{eqn:mode_product}
{\cal B}_{i_1\cdots i_{k-1} j i_{k+1} \cdots i_d} = \sum_{i_k=1}^{n_k} {\cal A}_{i_1\cdots i_{k-1} i_k i_{k+1} \cdots i_d} U_{j i_k},
\end{equation}
and ${\cal B} \in \mathbb{R}^{n_1 \times \cdots \times n_{k-1} \times n_k' \times n_{k+1} \times \cdots \times n_d}$. In particular, given a $d$-way tensor ${\cal A} \in \mathbb{R}^{n \times n \times \cdots \times n}$ and a vector ${\bf x} \in \mathbb{R}^n$, the multidimensional contraction, denoted by ${\cal A} {\bf x}^d$, is the scalar 
\begin{equation}\label{eqn:contraction}
{\cal A} {\bf x}^d  = {\cal A} \times_1 {\bf x}^\top \times_2 {\bf x}^\top \times_3 \cdots \times_d {\bf x}^\top,
\end{equation}
which is obtained as a homogeneous polynomial of ${\bf x} \in \mathbb{R}^n$ with degree $d$. The inner product of two same-sized tensors ${\cal A}, {\cal B} \in \mathbb{R}^{n_1 \times n_2 \times \cdots \times n_d}$ is the sum of the products of their entries, i.e.,
\begin{equation}\label{eqn:inner_product}
\langle {\cal A}, {\cal B} \rangle = \sum_{i_1=1}^{n_1} \sum_{i_2=1}^{n_2}\cdots \sum_{i_d=1}^{n_d} {\cal A}_{i_1 i_2 \cdots i_d} {\cal B}_{i_1 i_2 \cdots i_d} .
\end{equation}
The Frobenius norm of a tensor ${\cal A} \in \mathbb{R}^{n_1 \times n_2 \times \cdots \times n_d}$ is given by 
\begin{equation}\label{eqn:tensor_norm}
\| {\cal A} \|_F = \sqrt{\langle {\cal A}, {\cal A}\rangle }.
\end{equation}
The vectorization of a tensor ${\cal A} \in \mathbb{R}^{n_1 \times n_2 \times \cdots \times n_d}$ is denoted by $\textrm{vec}(\cal A)$ and maps the tensor element with indices $(i_1, i_2, \ldots, i_d)$ to the vector element with index $i$ where
$$
i=i_1 + (i_2-1) n_1 + \cdots + (i_d -1)\prod_{k=1}^{d-1} n_k .
$$
Given $d$ vectors ${\bf x}^{(i)} \in \mathbb{R}^{n_i}$, $i=1,2,\ldots, d$, their outer product is denoted by ${\bf x}^{(1)} \circ {\bf x}^{(2)} \circ \cdots \circ {\bf x}^{(d)}$, which is a tensor in $\mathbb{R}^{n_1 \times n_2 \times \cdots \times n_d}$ such that its entry with indices $(i_1, i_2, \ldots, i_d)$ is equal to the product of the corresponding vector elements, namely, $x^{(1)}_{i_1} x^{(2)}_{i_2}\cdots x^{(d)}_{i_d}$. It follows immediately that
\begin{equation}\label{eqn:vectorization}
\textrm{vec}({\bf x}^{(1)} \circ {\bf x}^{(2)} \circ \cdots \circ {\bf x}^{(d)}) = {\bf x}^{(d)} \otimes {\bf x}^{(d-1)} \otimes \cdots \otimes {\bf x}^{(1)} ,
\end{equation}
where the symbol ``$\otimes$'' denotes the Kronecker product.

We now illustrate how to represent a polynomial by using tensors. Denote by $\mathbb{R}[{\bf x}]$ the polynomial ring in $d$ variables ${\bf x}=(x_1, x_2, \ldots, x_d)^\top$ with coefficients in the field $\mathbb{R}$.

\begin{definition}\label{def:pure_power_poly}
Given a vector ${\bf \tilde{n}}=(\tilde{n}_1, \tilde{n}_2, \ldots, \tilde{n}_d) \in \mathbb{N}^d$, a polynomial $f \in \mathbb{R}[{\bf x}]$ with $d$ variables is called pure-power-$\bf \tilde{n}$ if the degree of $f$ is at most $\tilde{n}_i$ with respect to each variable $x_i$, $i=1, 2, \ldots, d$.
\end{definition}
\begin{example}
\label{ex:ex1}
The polynomial $f=4x_1+x_1^3-2x_1x_2x_3-7x_2x_3^2$ is a pure-power-$\bf \tilde{n}$ polynomial with ${\bf \tilde{n}}=(3,1,2)$.
\end{example}

The set of all pure-power-$\bf \tilde{n}$ polynomials with the degree vector ${\bf \tilde{n}}=(\tilde{n}_1, \tilde{n}_2, \ldots, \tilde{n}_d) \in \mathbb{N}^d$ is denoted by $\mathbb{R}[{\bf x}]_{\bf \tilde{n}}$. For any $f({\bf x}) \in \mathbb{R}[{\bf x}]_{\bf \tilde{n}}$, there are a total of $\prod_{k=1}^d(\tilde{n}_k + 1)$ distinct monomials 
$$
\prod_{k=1}^d x_k^{i_k-1}, \quad 1\leq i_k \leq \tilde{n}_k+1, \quad k=1,2,\ldots, d.
$$
For ${\bf x} = (x_1, x_2, \ldots, x_d)^\top \in \mathbb{R}^d$, denote by $\{ {\bf v}(x_k) \}_{k=1}^d$ the Vandermonde vectors
\begin{equation}\label{eqn:vander_vec}
{\bf v}(x_k) := (1, x_k, \ldots, x_k^{\tilde{n}_k})^\top \in \mathbb{R}^{\tilde{n}_k+1}.
\end{equation}
It follows that there is a one-to-one mapping between pure-power-$\bf \tilde{n}$ polynomials and tensors. To be specific, for any $f({\bf x}) \in \mathbb{R}[{\bf x}]_{\bf \tilde{n}}$, there exists a unique tensor ${\cal A} \in \mathbb{R}^{(\tilde{n}_1 +1) \times (\tilde{n}_2 +1) \times \cdots \times (\tilde{n}_d +1)}$ such that
\begin{equation}\label{eqn:poly2tensor}
f({\bf x}) = {\cal A} \times_1 {\bf v}(x_1)^\top \times_2 {\bf v}(x_2)^\top \times_3 \cdots \times_d {\bf v}(x_d)^\top. 
\end{equation}
\begin{example}
\label{ex:ex2}
We revisit the polynomial $f$ from Example \ref{ex:ex1} and illustrate its corresponding tensor representation. Since ${\bf \tilde{n}}=(3,1,2)$, we construct the Vandermonde vectors ${\bf v}(x_1)=(1,x_1,x_1^2,x_1^3), {\bf v}(x_2)=(1,x_2),{\bf v}(x_3)=(1,x_3,x_3^2)$. The nonzero entries of the corresponding $4 \times 2 \times 3$ tensor $\cal A$ are then ${\cal A}_{211}=4,{\cal A}_{411}=1,{\cal A}_{222}=-2,{\cal A}_{123}=-7$. The indices of the  tensor {\cal A} are easily found from grouping together corresponding indices of the Vandermonde vectors. For example, the tensor index $123$ corresponding with the monomial $x_2x_3^2$ is found from ${\bf v}(x_1)_1=1,{\bf v}(x_2)_2=x_2,{\bf v}(x_3)_3=x_3^2$.
\end{example}

\subsection{Tensor trains}
\label{sec:sub:tt}

It is well known that the number of tensor elements grows exponentially with the order $d$. Even when the dimensions are small, the storage cost for all elements is prohibitive for large $d$. The TT decomposition \cite{Ivan11a} gives an efficient way (in storage and computation) to overcome this so-called {\it curse of dimensionality}.

\begin{figure}[!t]
\centering
\includegraphics[width=3.5in]{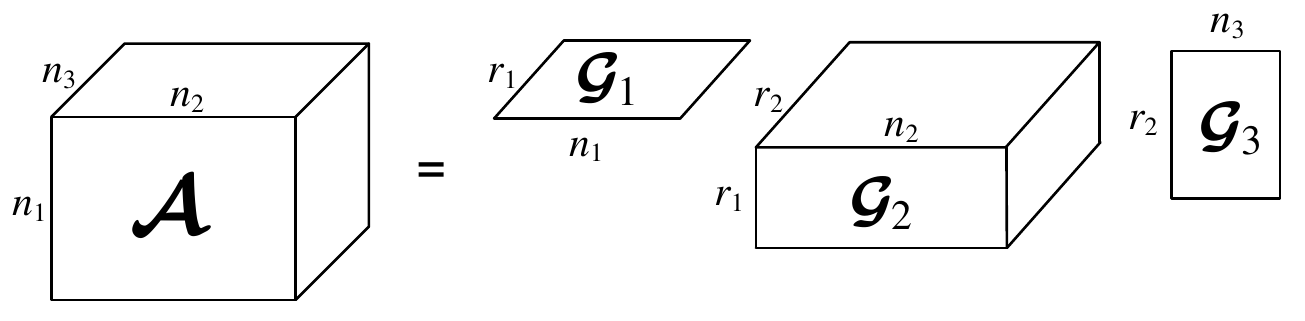}
\caption{The TT decomposition for a tensor in $\mathbb{R}^{n_1\times n_2 \times n_3}$.}
\label{fig:tt}
\end{figure}

The main idea of the TT decomposition is to re-express the entries of a tensor ${\cal A} \in \mathbb{R}^{n_1 \times n_2 \times \cdots \times n_d}$ as a product of matrices 
\begin{equation}\label{eqn:tt}
{\cal A}_{i_1 i_2 \cdots i_d} = {\cal G}_1(i_1) {\cal G}_2(i_2) \cdots {\cal G}_d(i_d) ,
\end{equation}
where ${\cal G}_k(i_k)$ is an $r_{k-1} \times r_k$ matrix for each index $i_k$, also called the TT-core. To turn the matrix-by-matrix product \eqref{eqn:tt} into a scalar, boundary conditions $r_0 = r_d = 1$ have to be introduced. The quantities $\{r_k\}_{k=0}^d$ are called the TT-ranks. Note that each core ${\cal G}_k$ is a third-order tensor with dimensions $r_{k-1}$, $n_k$ and $r_k$. The TT-decomposition for a tensor ${\cal A} \in \mathbb{R}^{n_1\times n_2 \times n_3}$ is illustrated in Fig.~\ref{fig:tt}. The most common way to convert a given tensor ${\cal A}$ into a TT would be the TT-SVD algorithm~\cite[p.~2301]{Ivan11a}.
\begin{example}TT-SVD algorithm~\cite[p.~2301]{Ivan11a}.
\label{ex:ex3}
Using the TT-SVD algorithm, we can convert the tensor ${\cal A}$ from Example \ref{ex:ex2} into a TT that consists of TT-cores ${\cal G}_1 \in \mathbb{R}^{1 \times 4 \times 3}, {\cal G}_2 \in \mathbb{R}^{3\times 2 \times 3}, {\cal G}_3 \in \mathbb{R}^{3 \times 3 \times 1}$.
\end{example}
Note that throughout this article, we will not need to use the TT-SVD algorithm. Instead, we will initialize the TT-cores randomly and iteratively update the cores one-by-one in an alternating fashion. It turns out that if all TT-ranks are bounded by $r$, the storage of the TT grows linearly with the order $d$ as $O(d n r^2 )$, where $n = \max \{ n_1, n_2, \ldots, n_d \}$.

\begin{proposition}[Theorem 2.1 of \cite{OT10}]\label{prop:upperbounds} For any tensor ${\cal A} \in \mathbb{R}^{n_1 \times n_2 \times \cdots \times n_d}$, there exists a TT-decomposition with TT-ranks 
$$
r_k \leq \min(\prod_{i=1}^{k} n_i,  \prod_{i=k+1}^d n_i), \quad k=1,2,\ldots, d-1.
$$
\end{proposition}

We also mention that the TT representation of a tensor is not unique. For instance, let $Q$ be an orthogonal matrix in $\mathbb{R}^{r_1 \times r_1}$, namely, $Q Q^\top = Q^\top Q= I_{r_1}$. Then the tensor $\cal A$ in (\ref{eqn:tt}) also has the TT-decomposition
\begin{equation}\label{eqn:tt2}
{\cal A}_{i_1 i_2 \cdots i_d} = {\cal G}_1'(i_1) {\cal G}_2'(i_2) \cdots {\cal G}_d(i_d) ,
\end{equation}
where 
$$
{\cal G}_1'(i_1) = {\cal G}_1(i_1) Q, \quad {\cal G}_2'(i_2) = Q^\top {\cal G}_2(i_2) .
$$
Numerical stability of our learning algorithms is guaranteed by keeping all the TT-cores left-orthogonal or right-orthogonal \cite{SO11}, which is achieved through a sequence of QR decompositions as explained in Section \ref{sec:LearningAlg}. 

\begin{definition}\label{def:let_right_orthogonal}
The $r_{k-1} \times n_k \times r_k$ core ${\cal G}_k$ is called left-orthogonal if 
$$
\sum_{i_k=1}^{n_k} {\cal G}_k(i_k)^\top {\cal G}_k(i_k) = I_{r_k}  ,
$$
and the $r_{k-1} \times n_k \times r_k$ core ${\cal G}_k$ is called right-orthogonal if 
$$
\sum_{i_k=1}^{n_k} {\cal G}_k(i_k) {\cal G}_k(i_k)^\top = I_{r_{k-1}}  .
$$
\end{definition}

As stated before, the structure of a TT also benefits the computation of the general multidimensional contraction:
\begin{equation}\label{eqn:general_contraction}
f = {\cal A} \times_1 ({\bf v}^{(1)})^\top \times_2 ({\bf v}^{(2)})^\top \times_3 \cdots \times_d ({\bf v}^{(d)})^\top ,
\end{equation}
where ${\cal A} \in \mathbb{R}^{n_1 \times n_2 \times \cdots \times n_d}$ and ${\bf v}^{(i)}=(v^{(i)}_1, v^{(i)}_2, \ldots, v^{(i)}_{n_i})^\top \in \mathbb{R}^{n_i}$, $i = 1,2,\ldots, d$.
If a tensor $\cal A$ is given in the TT-format~(\ref{eqn:tt}), then we have
\begin{equation}\label{eqn:tt_contraction}
f = \prod_{k=1}^d \sum_{i_k =1}^{n_k} v^{(k)}_{i_k} {\cal G}_k (i_k) .
\end{equation}
The described procedure for fast TT contraction is summarized in Algorithm~\ref{alg:ttv}. In order to simplify the analysis on the computational complexity of Algorithm \ref{alg:ttv}, we assume that $r_1=r_2=\cdots =r_{d-1}=r$ and $n_1=n_2=\cdots=n_d=n$. There are two required steps to compute the contraction of a TT with vector. First, we need to construct $d$ matrices $V^{(k)}$ by contracting the TT-cores ${\cal G}_k$ with the vectors ${\bf v}^{(k)}$. This operation is equivalent with $d$ matrix-vector products with a total computational cost of approximately $O(dr^2n)$ flops. Fortunately, the contraction of one TT-core is completely independent from the other contractions and hence can be done in parallel over $d$ processors, reducing the computational complexity to $O(r^2n)$ per processor. Maximal values for $r$ and $n$ in our experiments are 10 and 4, respectively, so that the contraction of one TT-core is approximately equivalent with the product of a $100\times 4$ matrix with a $4\times 1$ vector. The final step in Algorithm~\ref{alg:ttv} is the product of all matrices $V^{(k)}$ with a total computational complexity of $O(dr^2)$. If we again set $r=10$, $n=4$, then this final step in Algorithm~\ref{alg:ttv} is equivalent with the product of a $100\times 40$ matrix with a $40\times 1$ vector. For more basic operations implemented in the TT-format, such as tensor addition and computing the Frobenius norm, the reader is referred to \cite{Ivan11a}.
\begin{algorithm}
\caption{Fast TT contraction \cite{Ivan11a}}
\label{alg:ttv}
\begin{algorithmic}[1]
\REQUIRE{ Vectors ${\bf v}^{(k)} \in \mathbb{R}^{n_k}$, $k=1,2,\ldots, d$ and a tensor $\cal A$ in the TT-format with cores ${\cal G}_k$  }
\ENSURE{ The multidimensional contraction $f$ in (\ref{eqn:general_contraction}) }
\FOR{$k = 1:d $}
\STATE{ $V^{(k)} = \sum_{i_k =1}^{n_k} v^{(k)}_{i_k} {\cal G}_k (i_k)$  \hfill \%Computed in parallel} 
\ENDFOR
\STATE{ ${\bf f} := V^{(1)} $}
\FOR{$k = 2:d $ }
\STATE{ $ {\bf f} := {\bf f} V^{(k)}$}
\ENDFOR
\RETURN{$\bf f$}
\end{algorithmic}
\end{algorithm}

\section{TT Learning}
\label{sec:TTLearning}

\begin{figure}[!t]
\centering
\includegraphics[width=3.4in]{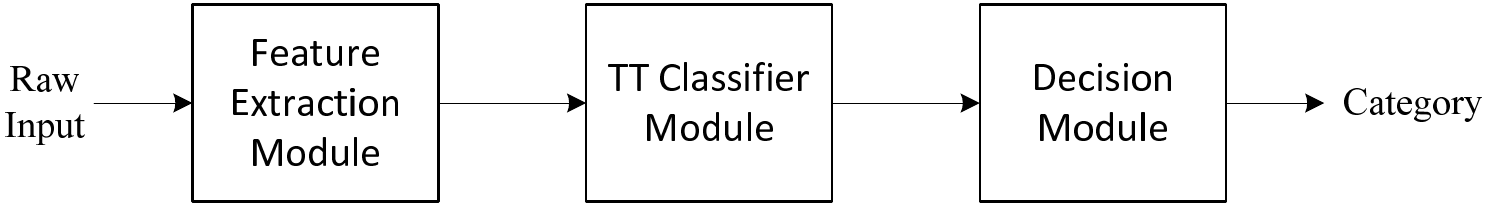}
\caption{Framework of TT learning.}
\label{fig_ttlearning}
\end{figure}

It is easy for us to recognize a face, understand spoken words, read handwritten characters and identify the gender of a person. Machines, however, make decisions based on data measured by a large number of sensors. In this section, we present the framework of TT learning. Like most pattern recognition systems \cite{TK08}, our TT learning method consists in dividing the system into three main modules, shown in Fig.~\ref{fig_ttlearning}.

The first module is called feature extraction, which is of paramount importance in any pattern classification problem. The goal of this module is to build features via transformations of the raw input, namely, the original data measured by a large number of sensors. The basic reasoning behind transform-based features is that an appropriately chosen transformation can exploit and remove information redundancies, which usually exist in the set of samples obtained by measuring devices. The set of features exhibit high {\it information packaging} properties compared with the original input samples. This means that most of the classification-related information is compressed into a relatively small number of features, leading to a reduction of the necessary feature space dimension. Feature extraction benefits training the classifier in terms of memory and computation, and also alleviates the problem of overfitting since we get rid of redundant information. To deal with the task of feature extraction, some linear or nonlinear transformation techniques are widely used. For example, the Karhunen-Lo\`eve transform, related to principal component analysis (PCA), is one popular method for feature generation and dimensionality reduction. A nonlinear kernel version of the classical PCA is called kernel PCA, which is an extension of PCA using kernel methods. The discrete Fourier transform (DFT) can be another good choice due to the fact that for many practical applications, most of the energy lies in the low-frequency components. Compared with PCA, the basis vectors in the DFT are fixed and problem-dependent, which leads to a low computational complexity.

The second module, the TT classifier, is the core of TT learning. The purpose of this module is to mark a new observation based on its features generated by the previous module. As will be discussed, the task of pattern classification can be divided into a sequence of binary classifications. For each particular binary classification, the TT classifier assigns to each new observation a score that indicates which class it belongs to. In order to construct a good classifier, we exploit the fact that we know the labels for each sample of a given dataset. The TT classifier is trained optimally with respect to an {\it optimality criterion}. In some ways, the TT classifier can be regarded as a kind of generalized linear classifier, it does a linear classification in a higher dimensional space generated by the items of a given pure-power polynomial. The local information is encoded by the products of features. In contrast to kernel-based SVM classifiers that work in the dual space, the TT classifier is able to work directly in the high dimensional space by exploiting the TT-format. Similar with the backpropagation algorithm for multilayer perceptrons, the structure of a TT allows for updating the cores in an alternating way. In the next section, we will describe the training of two TT classifiers through the optimization of two different loss functions.

The last module in Fig.~\ref{fig_ttlearning} is the decision module that decides which category a new observation belongs to. For binary classification, decisions are made according to the sign of the score assigned by the TT classifier, namely, the decision depends on the value of corresponding discriminant function. In an $m$-class problem, there are several strategies to decompose it into a sequence of binary classification problems. A straightforward extension is the {\it one-against-all}, where $m$ binary classification problems are involved. We seek to design discriminant functions $\{g_i({\bf x})\}_{i=1}^m$ so that $g_i({\bf x}) > g_j({\bf x})$, $\forall j \neq i$ if $\bf x$ belongs to the $i$th class. Classification is then achieved according to the rule:
\begin{center}
assign ${\bf x}$ to the $i$th class if $i=\text{argmax}_{k}\, g_k({\bf x})$. 
\end{center}
An alternative technique is the {\it one-against-one}, where we need to consider $m(m-1)/2$ pairs of classes. The decision is made on the basis of a majority vote. It means that each classifier casts one vote and the final class is the one with the most votes. When the number $m$ is too large, one can also apply the technique of binary coding. It turns out that only $\lceil \log_2 m \rceil$ classifiers are needed, where $\lceil\cdot \rceil$ is the ceiling operation. In this case, each class is represented by a unique binary code word of length $\lceil \log_2 m \rceil$. The decision is then made on the basis of minimal Hamming distance.

\section{Learning Algorithms}
\label{sec:LearningAlg}
For notational convenience, we define $n_k:=\tilde{n}_k+1$ and continue to use this notation for the remainder of the article. As stated before, TT classifiers are designed for binary classification. Given a set of $N$ training examples of the form $\{({\bf x}^{(j)}, y^{(j)})\}_{j=1}^N$ such that ${\bf x}^{(j)} \in \mathbb{R}^d$ is the feature vector of the $j$th example and $y^{(j)} \in \{-1, 1\}$ is the corresponding class label of ${\bf x}^{(j)}$. Let ${\bf \tilde{n}} = (\tilde{n}_1, \tilde{n}_2, \ldots, \tilde{n}_d)^\top \in \mathbb{N}^d$ be the degree vector. Each feature is then mapped to a higher dimensional space generated by all corresponding pure-power-$\bf \tilde{n}$ monomials through the mapping ${\cal T}: \mathbb{R}^d \rightarrow  \mathbb{R}^{n_1 \times n_2 \times \cdots \times n_d}$
\begin{equation}\label{eqn:kernel_map} 
{\cal T}({\bf x})_{i_1 i_2 \cdots i_d} = \prod_{k=1}^d x_k^{i_k-1}.
\end{equation}
For ${\bf x} =(x_1, x_2, \ldots, x_d)^\top \in \mathbb{R}^d$, let $\{ {\bf v}(x_k) \}_{k=1}^d$ be the Vandermonde vectors defined in (\ref{eqn:vander_vec}). Clearly, we have
\begin{equation}\label{eqn:rank1}
{\cal T}({\bf x}) = {\bf v}(x_1) \circ {\bf v}(x_2) \circ \cdots \circ {\bf v}(x_d). 
\end{equation}
This high-dimensional pure-power polynomial space benefits the learning task from the following aspects:
\begin{itemize}
\item all interactions between features are described by the monomials of pure-power polynomials;
\item the dimension of the tensor space grows exponentially with $d$, namely, $\prod_{k=1}^d n_k$, which increases the probability of separating all training examples linearly into two-classes; 
\item the one-to-one mapping between pure-power polynomials and tensors enables the use of tensor trains to lift the curse of dimensionality. 
\end{itemize}

With these preparations, our goal is to find a decision hyperplane to separate these two-class examples in the tensor space, also called the {\it generic feature space}. In other words, like the inductive learning described in \cite{SDDS12}, we try to find a tensor ${\cal A} \in \mathbb{R}^{n_1 \times n_2 \times \cdots \times n_d}$ such that 
\begin{align*}
y^{(j)}\langle {\cal T}({\bf x}^{(j)}), {\cal A} \rangle > 0, \quad j=1,2,\ldots, N.
\end{align*}
Note that the bias is absorbed into the first element of $\cal A$. Note that the learning problem can also be interpreted as finding a pure-power-${\bf \tilde{n}}$ polynomial $g({\bf x})$ such that 
$$
g({\bf x}^{(j)}) > 0, \quad \forall y^{(j)} =1,
$$ and 
$$
g({\bf x}^{(j)}) < 0, \quad \forall y^{(j)} =-1.
$$
Here we consider that the tensor $\cal A$ is expressed as a tensor train with cores $\{{\cal G}_k\}_{k=1}^d$. The main idea of the TT learning algorithms is to update the cores in an alternating way by optimizing an appropriate loss function. Prior to updating the TT-cores, the TT-ranks are fixed and a particular initial guess of $\{{\cal G}_k\}_{k=1}^d$ is made. The TT-ranks can be interpreted as tuning parameters, higher values will result in a better fit at the risk of overfitting. It is straightforward to extend our algorithms by means of the Density Matrix Renormalization Group (DMRG) method \cite{W92} such that the TT-ranks are updated adaptively. Each core is updated in the order
$$ {\cal G}_1 \rightarrow {\cal G}_2 \rightarrow \cdots \rightarrow {\cal G}_d \rightarrow {\cal G}_{d-1} \rightarrow \cdots \rightarrow {\cal G}_1 \rightarrow \cdots $$
until convergence, which is guaranteed under certain conditions as described in \cite{HRS12,RU13}. It turns out that updating one TT-core is equivalent with minimizing a loss function in a small number of variables, which can be done in a very efficient manner. The following lemma shows how the inner product $ \langle {\cal T}({\bf x}),  {\cal A} \rangle$ in the generic feature space is a linear function in any of the TT-cores ${\cal G}_k$.

\begin{lemma}\label{thm:core}
Given a vector ${\bf \tilde{n}} = (\tilde{n}_1, \tilde{n}_2, \ldots, \tilde{n}_d)^\top \in~\mathbb{N}^d$, let ${\cal T}$ be the mapping defined by (\ref{eqn:kernel_map}), and let $\cal A$ be a TT with cores ${\cal G}_k \in \mathbb{R}^{r_{k-1} \times n_k \times r_k}$, $k=1,2,\ldots, d$. For any ${\bf x} \in \mathbb{R}^d$ and $k=1,\ldots,d$, we have that 
\begin{align}
\langle {\cal T}({\bf x}),  {\cal A} \rangle = \left( {\bf q}_k({\bf x})^\top \otimes {\bf v}(x_k)^\top \otimes {\bf p}_k({\bf x}) \right)  {\rm vec}({\cal G}_k), 
\label{eq:innerprod}
\end{align}
where 
$$
{\bf p}_1({\bf x}) =1, \quad
\underset{k\geq 2} {{\bf p}_k({\bf x})} =\prod_{i=1}^{k-1}\left( {\cal G}_i \times_2 {\bf v}(x_i)^\top \right)  \in \mathbb{R}^{1\times r_{k-1}},
$$
and 
$$
\underset{k<d} {{\bf q}_k({\bf x})} =\prod_{i=k+1}^d \left( {\cal G}_i \times_2 {\bf v}(x_i)^\top \right)  \in \mathbb{R}^{r_{k} \times 1}, \quad {\bf q}_d({\bf x}) = 1.
$$
\end{lemma}
\noindent
{\it Proof.} By definition, we have
\begin{eqnarray*}
 \langle {\cal T}({\bf x}),  {\cal A} \rangle  & = &{\cal A} \times_1 {\bf v}(x_1)^\top \times_2 \cdots \times_d {\bf v}(x_d)^\top  \\
 & = & \left( {\cal G}_1 \times_2 {\bf v}(x_1)^\top \right) \cdots \left( {\cal G}_d \times_2 {\bf v}(x_d)^\top \right) \\
 & = & {\cal G}_k \times_1{\bf p}_k({\bf x}) \times_2 {\bf v}(x_k)^\top \times_3 {\bf q}_k({\bf x})^\top  \\
 & = & \left( {\bf q}_k({\bf x})^\top \otimes {\bf v}(x_k)^\top \otimes {\bf p}_k({\bf x}) \right)  \textrm{vec}({\cal G}_k)
\end{eqnarray*}
for any $k=1,2,\ldots, d$. This completes the proof.
\qed

\medskip
\begin{example}
In this example we illustrate the advantageous representation of a pure-power polynomial $f$ as a TT. Suppose we have a polynomial $f$ with $d=10$ and all degrees $\tilde{n}_i=9\, (i=1,\ldots,10)$. All coefficients of $f({\bf x})$ can then be stored into a 10-way tensor $10\times 10 \times \cdots \times 10$ tensor ${\cal A}$ such that the evaluation of $f$ in a particular ${\bf x}$ is given by \eqref{eqn:poly2tensor}. The TT-representation of $f$ consists of 10 TT-cores ${\cal G}_1,\ldots,{\cal G}_{10}$, with a storage complexity of $O(100r^2)$, where $r$ is the maximal TT-rank. This demonstrates the potential of the TT-representation in avoiding the curse of dimensionality when the TT-ranks are small.
\end{example}
\begin{example}
Next, we illustrate the expressions for ${\cal T({\bf x})},{\cal A}, {\bf v}(x_k),{\bf q}_k({\bf x}),{\bf p}_k({\bf x})$ for the following quadratic polynomial in two variables $f({\bf x})=1+3x_1-x_2-x_1^2+7x_1x_2+9x_2^2$. Since $d=2$ and $\tilde{n}_1=\tilde{n}_2=2$, both ${\cal T}$ and ${\cal A}$ are the following $3\times 3$ matrices
\begin{align*}
{\cal T({\bf x})} = \begin{pmatrix}1 & x_2 & x_2^2 \\ x_1 & x_1x_2 & x_1x_2^2 \\ x_1^2 & x_1^2x_2 & x_1^2x_2^2 \end{pmatrix},\, {\cal A} &= \begin{pmatrix}1 & -1 & 9 \\ 3 & 7 & 0 \\ -1 & 0 & 0 \end{pmatrix}.
\end{align*}
The TT-representation of ${\cal A}$ consists of a $1 \times 3 \times 3$ tensor ${\cal G}_1$ and a $3 \times 3 \times 1$ tensor ${\cal G}_2$. Suppose now that $k=2$ and we want to compute the evaluation of the polynomial $f$ in a particular ${\bf x}$, which is $\langle {\cal T({\bf x})},{\cal A} \rangle$. From Lemma \ref{thm:core} we then have that
\begin{align*}
\langle {\cal T({\bf x})},{\cal A} \rangle &= \left( {\bf q}_2({\bf x})^\top \otimes {\bf v}(x_2)^\top \otimes {\bf p}_2({\bf x}) \right)  {\rm vec}({\cal G}_2),
\end{align*}
with
\begin{align*}
{\bf q}_2({\bf x}) &= 1 \in \mathbb{R}, \\
{\bf v}(x_2) &= \begin{pmatrix} 1 & x_2 & x_2^2 \end{pmatrix}^\top \in \mathbb{R}^3,\\
{\bf p}_2({\bf x}) &= {\cal G}_1 \times_2 {\bf v}(x_1)^\top \in \mathbb{R}^{1\times 3},\\
{\bf v}(x_1) &= \begin{pmatrix} 1 & x_1 & x_1^2 \end{pmatrix}^\top \in \mathbb{R}^3.
\end{align*}
\end{example}

In what follows, we first present two learning algorithms based on different loss functions. These algorithms will learn the tensor ${\cal A}$ directly in the TT-representation from a given dataset. Two enhancements, namely, regularization for better accuracy and parallelization for higher speed will be described in the last two subsections.

\subsection{TT Learning by Least Squares}

Least squares estimation is the simplest and thus most common estimation method. In the generic feature space, we attempt to design a linear classifier so that its desired output is exactly $1$ or $-1$. However, we have to live with errors, that is, the true output will not always be equal to the desired one. The least squares estimator is then found from minimizing the following mean square error function 
\begin{equation}\label{eqn:ls_loss}
J({\cal A}) = \frac{1}{N} \sum_{j=1}^N \left( \langle T({\bf x}^{(j)}), {\cal A} \rangle - y^{(j)} \right)^2.
\end{equation}
We now show how updating a TT-core ${\cal G}_k$ is equivalent with solving a relatively small linear system. First, we define the $N \times r_{k-1}n_k r_k$ matrix 
\begin{align}\label{eqn:sub_matrix}
C_k= \left[ \begin{array}{c}
{\bf q}_k({\bf x}^{(1)})^\top \otimes {\bf v}(x_k^{(1)})^\top \otimes {\bf p}_k({\bf x}^{(1)})  \\
{\bf q}_k({\bf x}^{(2)})^\top \otimes {\bf v}(x_k^{(2)})^\top \otimes {\bf p}_k({\bf x}^{(2)})  \\
\vdots \\
{\bf q}_k({\bf x}^{(N)})^\top \otimes {\bf v}(x_k^{(N)})^\top \otimes {\bf p}_k({\bf x}^{(N)}) 
\end{array}
\right]
\end{align}
for any $k=1,2,\ldots, d$. The matrix $C_k$ is hence obtained from the concatenation of the row vectors ${\bf q}_k({\bf x})^\top \otimes {\bf v}(x_k)^\top \otimes {\bf p}_k({\bf x})$ from $\eqref{eq:innerprod}$ for $N$ samples ${\bf x}^{(1)},\ldots,{\bf x}^{(N)}$. It follows from Lemma~\ref{thm:core} that 
\begin{equation}\label{eqn:subls}
J({\cal A}) = \frac{1}{N} \| C_k\, \textrm{vec}({\cal G}_k) - {\bf y}\|^2
\end{equation}
where 
\begin{equation}\label{eqn:yvec}
{\bf y}=(y^{(1)}, y^{(2)}, \ldots, y^{(N)} )^\top \in \mathbb{R}^N.
\end{equation}
We have thus shown that updating the core ${\cal G}_k$ is equivalent with solving a least squares optimization problem in $r_{k-1}n_k r_k$ variables. Minimizing (\ref{eqn:subls}) with respect to ${\cal G}_k$ for any $k=1,\ldots,d$ results in solving the linear system
\begin{equation}\label{eqn:ls}
( C_k^\top C_k)\, \textrm{vec}({\cal G}_k) = C_k^\top {\bf y}.
\end{equation}
Supposing $r_1=r_2=\cdots =r_{d-1}=r$ and $n_1=n_2=\cdots = n_{d}=n$, then the computational complexity of solving \eqref{eqn:ls} is $O((r^2n)^3)$. For the maximal values of $r=10$ and $n=4$ in our experiments, this implies that we need to solve a linear system of order 400, which takes about 0.01 seconds using MATLAB on our desktop computer.

\subsection{TT Learning by Logistic Regression}

Since our goal is to find a hyperplane to separate two-class training examples in the generic feature space, we may not care about the particular value of the output. Indeed, only the sign of the output makes sense. This gives us the idea to decrease the number of sign differences as much as possible when updating the TT-cores, i.e., to minimize the number of misclassified examples. However, this model is discrete so that a difficult combinatorial optimization problem is involved. Instead, we try to find a suboptimal solution in the sense of minimizing a continuous cost function that penalizes misclassified examples. Here, we consider the logistic regression cost function. First, consider the standard sigmoid function
$$
\sigma(z) = \frac{1}{1+e^{-z}}, \quad z \in \mathbb{R},
$$
where the output always takes values between $0$ and $1$. An important property is that its derivative can be expressed by the function itself, i.e.,
\begin{align} 
\sigma'(z) = \sigma(z)(1-\sigma(z)) .
\label{eqn:derivativeproperty}
\end{align}
The logistic function for the $j$th example ${\bf x}^{(j)}$ is given by
\begin{equation}\label{eqn:log_fun}
h_{{\cal A}}({\bf x}^{(j)} ) := \sigma \left( \langle T({\bf x}^{(j)}), {\cal A} \rangle \right).
\end{equation}
We can also interpret the logistic function as the probability that the example ${\bf x}^{(j)}$ belongs to the class denoted by the label~$1$. The predicted label $\tilde{y}^{(j)}$ for ${\bf x}^{(j)}$ is then obtained according to the rule
$$
\left\{
\begin{aligned} 
& h_{{\cal A}}({\bf x}^{(j)} ) \geq 0.5  \Leftrightarrow \langle T({\bf x}^{(j)}), {\cal A} \rangle \geq 0 \rightarrow \tilde{y}^{(j)}=1, \\
& h_{{\cal A}}({\bf x}^{(j)} ) < 0.5  \Leftrightarrow \langle T({\bf x}^{(j)}), {\cal A} \rangle < 0 \rightarrow \tilde{y}^{(j)}=-1. 
\end{aligned}
\right.
$$
For a particular example ${\bf x}^{(j)}$, we define the cost function as
$$
\text{Cost}({\bf x}^{(j)}, {\cal A})=\left\{ 
\begin{aligned}
& -\ln \left( h_{\cal A}({\bf x}^{(j)}) \right)   &  & \text{if} \  y^{(j)}=1, \\ 
& -\ln \left( 1- h_{\cal A}({\bf x}^{(j)}) \right)   &  & \text{if} \  y^{(j)}=-1.  
\end{aligned}
\right.
$$
The goal now is to find a tensor ${\cal A}$ such that $h_{\cal A}({\bf x}^{(j)})$ is near $1$ if $y^{(j)}=1$ or near $0$ if $y^{(j)}=-1$. As a result, the logistic regression cost function for the whole training dataset is given by
\begin{equation}\label{eqn:log_loss}
\begin{aligned}
J({\cal A}) & = \frac{1}{N} \sum_{j=1}^N \text{Cost}({\bf x}^{(j)}, {\cal A}) \\
& = \frac{-1}{N} \sum_{j=1}^N \left[ \frac{1+y^{(j)}}{2} \ln\left( h_{\cal A}({\bf x}^{(j)}) \right)  + \right. \\ 
& \hspace{2.2cm} \left. \frac{1-y^{(j)}}{2} \ln\left( 1- h_{\cal A}({\bf x}^{(j)}) \right) \right] .
\end{aligned}
\end{equation}
It is important to note that~\eqref{eqn:log_loss} is convex though the sigmoid function is not. This guarantees that we can find the globally optimal solution instead of a local optimum.

From equation~(\ref{eqn:log_fun}) and Lemma~\ref{thm:core}, one can see that the function $J({\cal A})$ can also be regarded as a function of the core ${\cal G}_k$ since
$$
\langle T({\bf x}^{(j)}), {\cal A} \rangle = C_k(j, \ :)\, \textrm{vec}({\cal G}_k) 
$$
where $C_k(j, \ :)$ denote the $j$th row vector of $C_k$ defined in (\ref{eqn:sub_matrix}). It follows that updating the core ${\cal G}_k$ is equivalent with solving a convex optimization problem in $r_{k-1}n_kr_k$ variables. Let 
\begin{equation}\label{eqn:hvec}
{\bf h}_{\cal A} =\left( h_{\cal A}({\bf x}^{(1)}), h_{\cal A}({\bf x}^{(2)}), \ldots, h_{\cal A}({\bf x}^{(N)}) \right)^\top \in \mathbb{R}^N
\end{equation}
and $D_{\cal A}$ be the diagonal matrix in $\mathbb{R}^{N \times N}$ with the $j$th diagonal element given by
$ h_{\cal A}({\bf x}^{(j)}) \left( 1- h_{\cal A}({\bf x}^{(j)}) \right)$.
By using the property \eqref{eqn:derivativeproperty} one can derive the gradient and Hessian with respect to ${\cal G}_k$ as
\begin{equation}\label{eqn:log_gradient}
\nabla_{{\cal G}_k} J({\cal A}) = \frac{1}{N} C_k^\top \left({\bf h}_{\cal A} -  \frac{{\bf y} + {\bf 1}}{2} \right)
\end{equation}
and 
\begin{equation}\label{eqn:log_Hessian}
\nabla_{{\cal G}_k}^2 J({\cal A}) = \frac{1}{N} C_k^\top D_{\cal A} C_k ,
\end{equation}
respectively, where ${\bf y}$ is defined in (\ref{eqn:yvec}) and $\bf 1$ denotes the all-ones vector in $\mathbb{R}^N$. Although we do not have a closed-form solution to update the core ${\cal G}_k$, the gradient and Hessian allows us to find the solution by efficient iterative methods, e.g. Newton's method whose convergence is at least quadratic in a neighbourhood of the solution. A quasi-Newton method, like the Broyden-Fletcher-Goldfarb-Shanno (BFGS) algorithm, is another good choice if the inverse of the Hessian is difficult to compute.

\subsection{Regularization}
The cost functions \eqref{eqn:ls_loss} and \eqref{eqn:log_loss} of the two TT learning algorithms do not have any regularization term, which may result in overfitting and hence bad generalization properties of the obtained TT classifier. Next, we discuss how the addition of a regularization term to \eqref{eqn:ls_loss} and \eqref{eqn:log_loss} results in a small modification of the small optimization problem that needs to be solved when updating the TT-cores ${\cal G}_k$.

Consider the regularized optimization problem
\begin{equation}\label{eqn:regularization}
\tilde{J}({\cal A}) = J({\cal A}) + \gamma  R({\cal A} ) ,
\end{equation}
where $J({\cal A})$ is given by \eqref{eqn:ls_loss} or \eqref{eqn:log_loss}, $\gamma$ is a parameter that balances the loss function and the regularization term. Here we use the Tikhonov regularization, namely,
\begin{equation}\label{eqn:Tik_term}
R({\cal A}) = \frac{1}{2} \langle {\cal A}, {\cal A} \rangle .
\end{equation}
Thanks to the TT structure, the gradient of $R({\cal A})$ with respect to the TT-core ${\cal G}_k$ can be equivalently rewritten as a linear transformation of $\textrm{vec}({\cal G}_k)$. In other words, there is a matrix $D_k \in \mathbb{R}^{r_{k-1}n_kr_k \times r_{k-1}n_kr_k}$ determined by the cores $\{ {\cal G}_j \}_{j \neq k}$ such that $\nabla_{{\cal G}_k} R({\cal A})=D_k \textrm{vec}({\cal G}_k)$. See Appendix~\ref{app:grad_Tiknorm} for more details. It follows that 
$$
\nabla_{{\cal G}_k} \tilde{J}({\cal A}) = \nabla_{{\cal G}_k} J({\cal A}) + \gamma D_k \textrm{vec} ({\cal G}_k) 
$$
and
$$
\nabla_{{\cal G}_k}^2 \tilde{J}({\cal A}) = \nabla_{{\cal G}_k}^2 J({\cal A}) + \gamma D_k. 
$$
These small modifications lead to small changes when updating the core ${\cal G}_k$. For instance, the first-order condition of \eqref{eqn:regularization} for the least squares model results in solving the modified linear system
\begin{equation}\label{eqn:ls_reg}
\left( C_k^\top C_k + \frac{N}{2} \gamma D_k \right) \textrm{vec}({\cal G}_k) = C_k^\top {\bf y},
\end{equation}
when compared with the original linear system \eqref{eqn:ls}.

\subsection{Orthogonalization and Parallelization}

The matrix $C_k$ from (\ref{eqn:sub_matrix}) needs to be reconstructed for each TT-core ${\cal G}_k$ during the execution of the two TT learning algorithms. Fortunately, this can be done efficiently by exploiting the TT structure. In particular, after updating the core ${\cal G}_k$ in the left-to-right sweep, the new row vectors $\{ {\bf p}_{k+1}({\bf x}^{(j)}) \}_{j=1}^N$ to construct the next matrix $C_{k+1}$ can be easily computed from
$$
{\bf p}_{k+1}({\bf x}^{(j)}) = {\cal G}_k \times_1 {\bf p}_{k}({\bf x}^{(j)}) \times_2 {\bf v}(x^{(j)}_k)^\top.
$$
Similarly, in the right-to-left sweep, the new column vectors $\{ {\bf q}_{k-1}({\bf x}^{(j)}) \}_{j=1}^N$ to construct the next matrix $C_{k-1}$ can be easily computed from
$$
{\bf q}_{k-1}({\bf x}^{(j)}) = {\cal G}_k \times_2 {\bf v} (x^{(j)}_k)^\top \times_3 {\bf q}_{k}({\bf x}^{(j)})^\top.
$$

To make the learning algorithms numerically stable, the techniques of orthogonalization are also applied. The main idea is to make sure that before updating the core ${\cal G}_k$, the cores ${\cal G}_1, \ldots, {\cal G}_{k-1}$ are left-orthogonal and the cores ${\cal G}_{k+1}, \ldots, {\cal G}_d$ are right-orthogonal by a sequence of QR decompositions. In this way, the condition number of the constructed matrix $C_k$ is upper bounded so that the subproblem is well-posed. After updating the core ${\cal G}_k$, we orthogonalize it with an extra QR decomposition, and absorb the upper triangular matrix into the next core (depending on the direction of updating). More details on the orthogonalization step can be found in~\cite{HRS12}.

Another computational challenge is the potentially large size $N$ of the training dataset. Luckily, the dimension of the optimization problem when updating ${\cal G}_k$ in the TT learning algorithms is $r_{k-1}(n_k+1)r_k$, which is much smaller and independent from $N$. We only need to compute the products $C_k^\top C_k$, $C_k^\top {\bf y}$, $C_k^\top {\bf h}_{\cal A}$ and $C_k^\top D_{\cal A} C_k$ in (\ref{eqn:ls}), (\ref{eqn:log_gradient}) and (\ref{eqn:log_Hessian}). These computations are easily done in parallel. Specifically, given a proper partition $\{N_l\}_{l=1}^L$ satisfying $\sum_{l=1}^L N_l = N$, we divide the large matrix $C_k$ into several blocks, namely,
$$
C_k = \left[ \begin{array}{c}
C_k^{(1)}   \\
C_k^{(2)}   \\
\vdots \\
C_k^{(L)} 
\end{array}
\right]  \in \mathbb{R}^{N \times r_{k-1}(n_k+1) r_k},
$$
where $C_k^{(l)} \in \mathbb{R}^{N_l \times r_{k-1}n_k r_k}$, $l=1,2,\ldots, L$. Then, for example, the product $C_k^\top D_{\cal A} C_k$ can be computed by
$$
C_k^\top D_{\cal A} C_k = \sum_{l=1}^L (C_k^{(l)})^\top D_{\cal A}^{(l)} C_k^{(l)},
$$
where $D_{\cal A}^{(l)}$ denotes the corresponding diagonal block. Each term in the summation on the right-hand side of the above equation can be computed over $L$ distributed cores, with a computational complexity of $O(r^4n_k^2N_l)$ for each core, supposing $r_{k-1}=r_k=r$. The other matrix products can also be computed in a similar way.

We summarize our learning algorithms in Algorithm \ref{alg:TTLearning}. The two most computationally expensive steps are lines 5 and 7. As we mentioned before, solving \eqref{eqn:regularization} takes approximately $O((r^2n)^3)$ flops. If the QR decomposition of line 7 is computed through Householder transformations, then the computational complexity is approximately $O(r^3n^2)$ flops. For the maximal values of $n=4$ and $r=10$ in our experiments, this amounts to solving computing the inverse and the QR factorization of a $400\times 400$ matrix. Note that based on the decision strategy, an $m$-class problem is decomposed into a sequence of two-class problems whose TT classifiers can be trained in parallel. 

\begin{algorithm}
\caption{Tensor Train Learning Algorithm}
\label{alg:TTLearning}
\begin{algorithmic}[1]
\REQUIRE{ Training dataset of pairs $\{({\bf x}^{(j)}, y^{(j)} )\}_{j=1}^N$, TT-ranks $\{r_k\}_{k=1}^{d-1}$, degree vector ${\bf \tilde{n}}= (\tilde{n}_1, \tilde{n}_2, \ldots, \tilde{n}_d)^\top \in \mathbb{N}^d$ and regularization parameter $\gamma$ }
\ENSURE{ Tensor $\cal A$ in TT format with cores $\{{\cal G}_k\}_{k=1}^d$ }
\STATE{Initialize right orthogonal cores $\{{\cal G}_k\}_{k=1}^d$ of prescribed ranks }
\WHILE{ termination condition is not satisfied }
\STATE  \%Left-to-right sweep
\FOR{$k = 1,2, \ldots, d-1 $} 
\STATE{${\cal G}_k^* \leftarrow $ find the minimal solution  of the regularized optimization problem (\ref{eqn:regularization}) with respect to ${\cal G}_k$}
\STATE{$U_k \leftarrow reshape({\cal G}_k^*, r_{k-1}n_k, r_k)$  }
\STATE{$[Q, R] \leftarrow$ compute QR decomposition of $U_k$ }
\STATE{${\cal G}_k \leftarrow reshape(Q, r_{k-1}, n_k, r_k)$ }
\STATE{$V_{k+1} \leftarrow R * reshape({\cal G}_{k+1}, r_k, n_{k+1}, r_{k+1})$ }
\STATE{${\cal G}_{k+1} \leftarrow reshape(V_{k+1}, r_k, n_{k+1}, r_{k+1})$ }
\ENDFOR
\STATE{Perform the right-to-left sweep}
\ENDWHILE
\end{algorithmic}
\end{algorithm}

We end this section with the following remarks:
\begin{itemize}
\item Other loss functions can also be used in the framework of TT learning provided that there exists an efficient way to solve the corresponding subproblems. 
\item The DMRG method~\cite{W92} can also be used to update the cores. This involves updating two cores at a time so that the TT-ranks are adaptively determined by means of a singular value decomposition (SVD). This may give better performance at the cost of a higher computational complexity. It also removes the need to fix the TT-ranks a priori.
\item The local linear convergence of Algorithm \ref{alg:TTLearning} has been established in \cite{HRS12,RU13} under certain conditions. In particular, if the TT-ranks are correctly estimated for convex optimization problems, then the obtained solution is guaranteed to be the global optimum. When choosing the TT-ranks, one should keep the upper bounds of the TT-ranks from Proposition \ref{prop:upperbounds} in mind.
\end{itemize}

\section{Experiments}
\label{sec:experiments}

In this section, we test our TT learning algorithms and compare their performance with LS-SVMs with polynomial kernels on two popular digit recognition datasets: USPS and MNIST. All our algorithms were implemented in MATLAB Version R2016a, which can be freely downloaded from \url{https://github.com/kbatseli/TTClassifier}. We compare our TT-polynomial classifiers with a polynomial classifier based on LS-SVMs with a polynomial kernel. The LS-SVM-polynomial classifier was trained with the MATLAB LS-SVMlab toolbox, which can be freely downloaded from \url{http://www.esat.kuleuven.be/sista/lssvmlab/}. The numerical experiments were done on a desktop PC with an Intel i5 quad-core processor running at 3.3GHz and 16GB of RAM.

The US Postal Service (USPS) database\footnote{The USPS database is downloaded from \\ \url{http://statweb.stanford.edu/~tibs/ElemStatLearn/data.html}}contains 9298 handwritten digits, including 7291 for training and 2007 for testing. Each digit is a 16$\times$16 grayscale image. It is known that the USPS test set is rather difficult and the human error rate is $2.5\%$.  The Modified NIST (MNIST) database\footnote{The MNIST database is downloaded from \\ \url{http://yann.lecun.com/exdb/mnist/}} of handwritten digits has a training set of 60,000 examples, and a test set of 10,000 examples. It is a subset of a larger set available from NIST. The digits have been size-normalized and centered in a 28$\times$28 image. The description of these two databases is summarized in Table~\ref{Data_description}.

\begin{table}[!t]
\caption{Dataset Description}
\label{Data_description}
\centering
\begin{tabular}{cccc}
\hline  \vspace{-2.5mm}   \\   
        & Image size     & Training size  & Test size   \\ 
\hline  \vspace{-2mm}   \\ 
USPS    & $16 \times 16$ & 7291           & 2007        \\
MNIST   & $28 \times 28$ & 60000          & 10000       \\
\hline
\end{tabular}
\end{table} 

Before extracting features of the handwritten digits, we first execute the pre-process of deskewing which is the process of straightening an image that has been scanned or written crookedly. By choosing a varying number $d$, the corresponding feature vectors are then extracted from a pre-trained CNN model 1-20-P-100-P-$d$-10, which represents a net with an input images of size 28$\times$28, a convolutional layer with 20 maps and 5$\times$5 filters, a max-pooling layer over non-overlapping regions of size 2$\times$2, a convolutional layer with 100 maps and 5$\times$5 filters, a max-pooling layer over non-overlapping regions of size 2$\times$2, a convolutional layer with $d$ maps and 4$\times$4 filters, and a fully connected output layer with 10 neurons. As the variants of the well-known CNN model LeNet-5 \cite{LBBH98}, these CNN models have been trained well on the MNIST database. For an input 28$\times$28 image, we get a feature vector of length $d$ from the third hidden layer. Note that for USPS database, we must resize image data to the size of image input layer. We mention that these techniques of deskewing and feature extraction using pre-trained CNN model are widely used in the literature \cite{LBBH98,DS02,LSB07}.

For the decision module, we adopt the one-against-all decision strategy where ten TT classifiers are trained to separate each digit from all the others. In the implementation of Algorithm~\ref{alg:TTLearning}, we normalize each initial core such that its Frobenius norm is equal to one. The degree vector is given by $\tilde{n}_1=\cdots =\tilde{n}_d=\tilde{n}$. The TT-ranks are upper bounded by $r_{\max}$. The values of $d,\tilde{n},r_{\max}$ were chosen to minimize the test error rate and to ensure that each of the subproblems to update the TT-cores ${\cal G}_k$ could be solved in a reasonable time. The dimension of each subproblem is at most $n\,r_{\max}^2$. For example, in the USPS case, we first fixed the values of $n$ and $r_{\max}$. We then fixed the value of $d$ and incremented $n$ and $r_{\max}$ to see whether this resulted in a better test error rate.
We use the optimality criterion
$$
\frac{| \tilde{J}({\cal A}^+) - \tilde{J}(\cal A) |}{| \tilde{J}(\cal A) |}  \leq 10^{-2} ,
$$
where ${\cal A}^+$ is the updated tensor from tensor $\cal A$ after one sweep. And the maximum number of sweeps is 4, namely, $4(d-1)$ iterations through the entire training data are performed for each session. To simplify notations, we use ``TTLS'' and ``TTLR'' to denote the TT learning algorithms based on minimizing loss functions (\ref{eqn:ls_loss}) and (\ref{eqn:log_loss}), respectively. For these two models, the regularization parameter $\gamma$ is determined by the technique of 10-fold cross-validation. In other words, we randomly assign the training data to ten sets of equal size. The parameter $\gamma$ is chosen so that the mean over all test errors is minimal.



\begin{figure}[!t]
\centering
\includegraphics[width=3.7in]{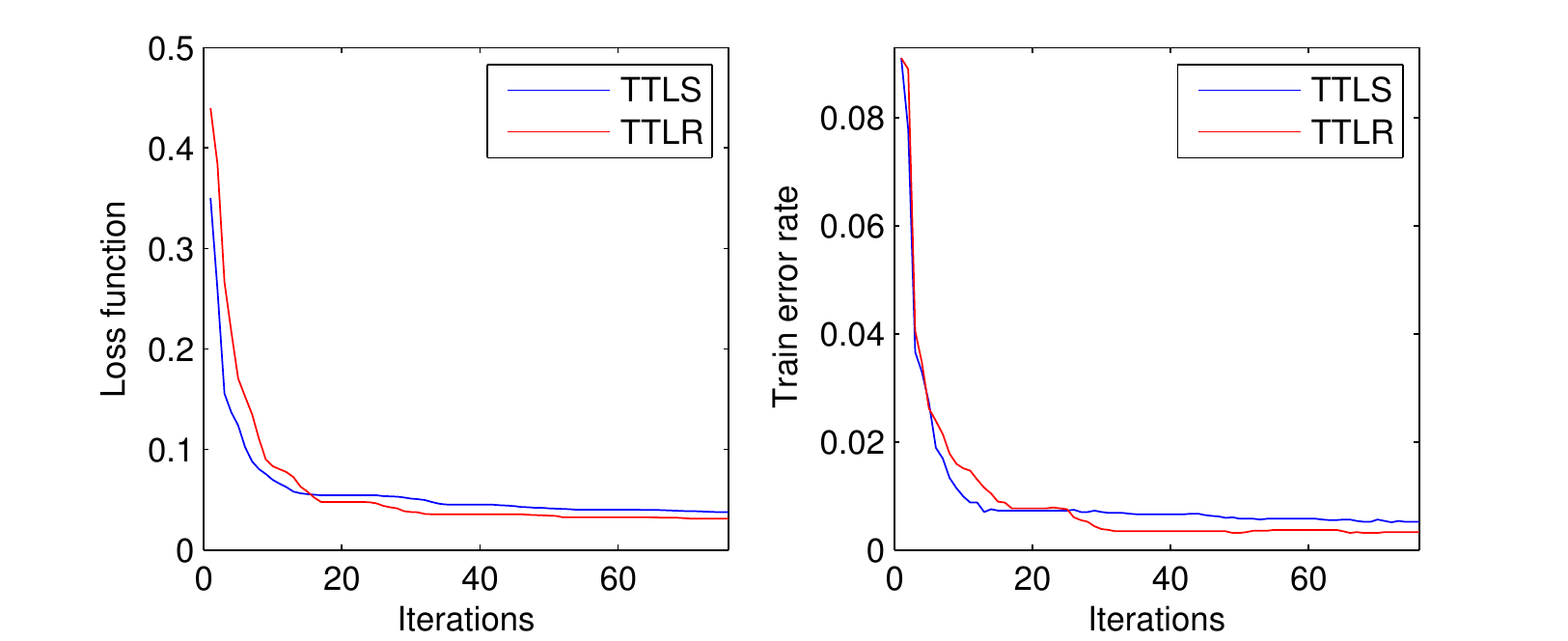}
\caption{The convergence of TT learning algorithms.}
\label{fig:tt_convergence}
\end{figure}

The numerical results for USPS database and MNIST database are reported in Tables~\ref{Result_USPS} and \ref{Result_MNIST}, respectively. The monotonic decrease is always seen when training the ten TT classifiers. Fig.~\ref{fig:tt_convergence} shows the convergence of both TT learning algorithms on the USPS data for the case $d=20, \tilde{n}=1, r_{\max}=8$ when training the classifier for the character $``6"$. In addition, we also trained a polynomial classifier using LS-SVMs with polynomial kernels on these two databases. Using the basic LS-SVM scheme, a training error rate of $0$ and a test error rate of $8.37\%$ were obtained for the USPS dataset after more than three and a half hours of computation. This runtime includes the time required to tune the tuning parameters via 10-fold cross-validation. When using an RBF kernel with the LS-SVM, it is possible to attain a test error of 2.14\%~\cite{Suykens2017}, but then the classifier is not polynomial anymore. The MNIST dataset resulted in consistent out-of-memory errors, which is to be expected as the basic SVM scheme is not intended for large data sets. We would also like to point out that a test error of $1.1\%$ is reported on the MNIST website for a polynomial classifier of degree 4 that uses conventional SVMs and deskewing. 

\begin{table*}[!t]
\renewcommand{\arraystretch}{1.3}
\caption{Numerical results for dataset USPS}
\label{Result_USPS}
\centering
\begin{tabular}{cccrrrcrrr}
\hline
& & & \multicolumn{3}{c}{TTLS} & & \multicolumn{3}{c}{TTLR} \\
\cline{4-6} \cline{8-10}  
$d$ & $\tilde{n}$ & $r_{\max}$ & Train error & Test error & Time(s) & & Train error & Test error & Time(s) \\
\hline
20 & 1 & 8  & 0.58\% & 4.33\% & 1.10$\times$10 & &  0.45\% & 4.33\% &  4.87$\times$10   \\
25 & 1 & 8  & 0.47\% & 4.04\% & 1.57$\times$10 & &  0.27\% & 3.99\% &  6.49$\times$10   \\
30 & 1 & 8  & 0.49\% & 3.84\% & 1.83$\times$10 & &  0.30\% & 3.99\% &  8.18$\times$10   \\
35 & 1 & 8  & 0.38\% & 4.04\% & 2.21$\times$10 & &  0.26\% & 4.24\% &  9.77$\times$10   \\
40 & 1 & 8  & 0.34\% & 3.89\% & 2.59$\times$10 & &  0.14\% & 3.74\% & 11.39$\times$10   \\
20 & 2 & 8  & 0.49\% & 4.29\% & 2.04$\times$10 & &  0.55\% & 4.24\% &  8.29$\times$10   \\
25 & 2 & 8  & 0.38\% & 4.29\% & 2.68$\times$10 & &  0.29\% & 4.14\% & 10.82$\times$10   \\
30 & 2 & 8  & 0.51\% & 4.14\% & 3.28$\times$10 & &  0.47\% & 3.99\% & 13.34$\times$10   \\
35 & 2 & 8  & 0.45\% & 4.38\% & 3.87$\times$10 & &  0.32\% & 4.38\% & 15.88$\times$10   \\
40 & 2 & 8  & 0.34\% & 4.09\% & 4.48$\times$10 & &  0.21\% & 3.74\% & 18.39$\times$10   \\
20 & 3 & 8  & 0.59\% & 4.58\% & 3.26$\times$10 & &  0.64\% & 4.33\% & 11.47$\times$10   \\
25 & 3 & 8  & 0.49\% & 4.53\% & 4.07$\times$10 & &  0.56\% & 4.14\% & 14.98$\times$10   \\
30 & 3 & 8  & 0.63\% & 4.43\% & 4.93$\times$10 & &  0.51\% & 4.09\% & 18.35$\times$10   \\
20 & 1 & 10 & 0.56\% & 4.33\% & 1.91$\times$10 & &  0.44\% & 4.33\% & 17.69$\times$10   \\
25 & 1 & 10 & 0.41\% & 3.94\% & 2.48$\times$10 & &  0.27\% & 3.94\% & 24.72$\times$10   \\
30 & 1 & 10 & 0.51\% & 3.84\% & 3.09$\times$10 & &  0.25\% & 4.04\% & 31.14$\times$10   \\
35 & 1 & 10 & 0.34\% & 3.94\% & 3.71$\times$10 & &  0.22\% & 4.29\% & 38.17$\times$10   \\
40 & 1 & 10 & 0.44\% & 3.84\% & 4.36$\times$10 & &  0.33\% & 3.84\% & 45.24$\times$10   \\
\hline
\end{tabular}
\end{table*}

\begin{table*}[!t]
\renewcommand{\arraystretch}{1.3}
\caption{Numerical results for dataset MNIST}
\label{Result_MNIST}
\centering
\begin{tabular}{cccrrrcrrr}
\hline
& & & \multicolumn{3}{c}{TTLS} & & \multicolumn{3}{c}{TTLR} \\
\cline{4-6} \cline{8-10}  
$d$ & $\tilde{n}$ & $r_{\max}$ & Train error & Test error & Time(s) & & Train error & Test error & Time(s) \\
\hline
20 & 1 & 8  & 0.16\% & 0.96\% &  7.14$\times$10 & & 0.12\% & 0.99\% & 17.52$\times$10   \\
25 & 1 & 8  & 0.15\% & 0.97\% &  9.38$\times$10 & & 0.09\% & 0.98\% & 22.96$\times$10   \\
30 & 1 & 8  & 0.13\% & 0.86\% & 11.73$\times$10 & & 0.10\% & 0.86\% & 28.96$\times$10   \\
35 & 1 & 8  & 0.19\% & 0.81\% & 13.95$\times$10 & & 0.03\% & 0.82\% & 34.55$\times$10   \\
40 & 1 & 8  & 0.21\% & 0.94\% & 16.22$\times$10 & & 0.07\% & 0.83\% & 40.31$\times$10   \\
20 & 2 & 8  & 0.19\% & 1.03\% & 12.08$\times$10 & & 0.14\% & 1.01\% & 29.45$\times$10   \\
25 & 2 & 8  & 0.13\% & 1.00\% & 15.72$\times$10 & & 0.13\% & 0.98\% & 38.32$\times$10   \\
30 & 2 & 8  & 0.19\% & 0.88\% & 18.94$\times$10 & & 0.10\% & 0.91\% & 46.77$\times$10   \\
35 & 2 & 8  & 0.17\% & 0.86\% & 23.11$\times$10 & & 0.13\% & 0.83\% & 56.54$\times$10   \\
40 & 2 & 8  & 0.17\% & 0.90\% & 25.67$\times$10 & & 0.17\% & 0.88\% & 64.68$\times$10   \\
20 & 3 & 8  & 0.20\% & 1.10\% & 17.45$\times$10 & & 0.19\% & 1.00\% & 42.24$\times$10   \\
30 & 3 & 8  & 0.22\% & 1.02\% & 27.21$\times$10 & & 0.21\% & 0.90\% & 65.89$\times$10   \\
40 & 3 & 8  & 0.21\% & 0.97\% & 36.48$\times$10 & & 0.07\% & 0.89\% & 90.37$\times$10   \\
20 & 1 & 10 & 0.20\% & 0.99\% & 10.14$\times$10 & & 0.19\% & 0.98\% & 36.66$\times$10   \\
30 & 1 & 10 & 0.16\% & 0.88\% & 17.15$\times$10 & & 0.17\% & 0.90\% & 62.24$\times$10   \\
40 & 1 & 10 & 0.17\% & 0.88\% & 24.05$\times$10 & & 0.13\% & 0.89\% & 83.62$\times$10   \\
40 & 2 & 10 & 0.13\% & 0.93\% & 39.61$\times$10 & & 0.14\% & 0.92\% & 144.1$\times$10   \\
40 & 3 & 10 & 0.18\% & 0.94\% & 59.66$\times$10 & & 0.10\% & 0.90\% & 207.7$\times$10   \\
40 & 4 & 10 & 0.24\% & 1.02\% & 78.83$\times$10 & & 0.14\% & 0.91\% & 264.2$\times$10   \\
\hline
\end{tabular}
\end{table*}

\section{Conclusion}
\label{sec:conclusion}

This paper presents the framework of TT learning for pattern classification. For the first time, TTs are used to represent polynomial classifiers, enabling the learning algorithms to work directly in the high-dimensional feature space. Two efficient learning algorithms are proposed based on different loss functions. The numerical experiments show that each TT classifier is trained in up to several minutes with competitive test errors. When compared with other polynomial classifiers, the proposed learning algorithms can be easily parallelized and have considerable advantage on storage and computation time. We also mention that these results can be improved by adding virtual examples \cite{DS02}. Future improvements are the implementation of on-line learning algorithms, together with the extension of the binary TT classifier to the multi-class case.


%


\appendices
\section{}
\label{app:grad_Tiknorm}

Given the degree vector ${\bf \tilde{n}}=(\tilde{n}_1, \tilde{n}_2, \ldots, \tilde{n}_d) \in \mathbb{N}^d$, let ${\cal A} \in \mathbb{R}^{n_1 \times n_2 \times \cdots \times n_d}$ be the tensor in TT format with cores ${\cal G}_k  \in \mathbb{R}^{r_{k-1} \times n_k \times r_k}$, $k=1, 2, \ldots, d$. To investigate the gradient of $R({\cal A})$ in~\eqref{eqn:Tik_term} with respect to the TT-core ${\cal G}_k$, we give a small variation $\epsilon$ to the $i$th element of $\textrm{vec}({\cal G}_k)$, resulting in a new tensor ${\cal A}_{\epsilon}$ given by
$$
{\cal A}_{\epsilon} = {\cal A} + \epsilon {\cal I}^{(k)}_i ,
$$
where $1\leq i \leq r_{k-1}n_kr_k$ and ${\cal I}^{(k)}_i$ is the tensor which has the same TT-cores with $\cal A$ except that the vectorization of the core ${\cal G}_k$ is replaced by the unit vector in $\mathbb{R}^{r_{k-1}n_kr_k}$ with the $i$th element equal to $1$ and $0$ otherwise. Then we have
\begin{equation}\label{eqn:Tik_grad1}
\left[\nabla_{{\cal G}_k} R({\cal A}) \right]_i = \lim_{\epsilon \rightarrow 0} \frac{R({\cal A}_\epsilon) - R(A)}{\epsilon} = \langle {\cal A}, {\cal I}^{(k)}_i \rangle . 
\end{equation}

On the other hand, by the definition of vectorization, the $i$th element of $\textrm{vec}({\cal G}_k) \in \mathbb{R}^{r_{k-1}n_kr_k}$ is mapped from the tensor element of ${\cal G}_k  \in \mathbb{R}^{r_{k-1} \times n_k \times r_k}$ with indices $(\alpha_{k-1}, j_k, \alpha_k)$ satisfying
$$
i=\alpha_{k-1} + (j_k-1) r_{k-1} + (\alpha_k -1) r_{k-1} n_k ,
$$
where $1\leq \alpha_{k-1} \leq r_{k-1}$, $1\leq j_k \leq n_k$ and $1\leq \alpha_k \leq r_k$. Denote by $E^{(\alpha_{k-1}, \alpha_k)}$ the matrix in $\mathbb{R}^{r_{k-1} \times r_k}$ such that the element with index $(\alpha_{k-1}, \alpha_k)$ equal to $1$ and $0$ otherwise. By simple computation, one can obtain that  
\begin{align}\label{eqn:Tik_grad2}
\langle {\cal A}, {\cal I}^{(k)}_i \rangle  &= \sum_{i_1, i_2, \ldots i_d} {\cal A}_{i_1 i_2 \cdots i_d} ({\cal I}^{k}_i)_{i_1 i_2 \cdots i_d}  \nonumber \\ 
&= {\bf a}_k \left( E^{(\alpha_{k-1}, \alpha_k)} \otimes {\cal G}_k(j_k) \right)  {\bf b}_k,  
\end{align}
where
\begin{equation}\label{eqn:Tik_grad_left}
{\bf a}_k =  \prod_{l=1}^{k-1} \sum_{i_l=1}^{n_l} \big[ {\cal G}_l(i_l) \otimes {\cal G}_l(i_l) \big] \in \mathbb{R}^{1 \times r_{k-1}^2}
\end{equation}
and
\begin{equation}\label{eqn:Tik_grad_right}
{\bf b}_k =  \prod_{l=k+1}^{d} \sum_{i_l=1}^{n_l} \big[ {\cal G}_l(i_l) \otimes {\cal G}_l(i_l) \big] \in \mathbb{R}^{r_{k}^2 \times 1} .
\end{equation}
Let ${\bf a}_k^{(1)}, {\bf a}_k^{(2)}, \ldots, {\bf a}_k^{(r_{k-1})} \in \mathbb{R}^{1 \times r_{k-1}}$ be the row vectors such that 
$$
{\bf a}_k = ({\bf a}_k^{(1)}, {\bf a}_k^{(2)}, \ldots, {\bf a}_k^{(r_{k-1})}) \in \mathbb{R}^{1 \times r_{k-1}^2} ,
$$
and let ${\bf b}_k^{(1)}, {\bf b}_k^{(2)}, \ldots, {\bf b}_k^{(r_{k})} \in \mathbb{R}^{r_{k} \times 1}$ be the column vectors such that 
$$
{\bf b}_k = \left[ \begin{array}{c}
{\bf b}_k^{(1)}   \\
{\bf b}_k^{(2)}   \\
\vdots \\
{\bf b}_k^{(r_k)} 
\end{array}
\right]  \in \mathbb{R}^{ r_{k}^2 \times 1},
$$
Combining \eqref{eqn:Tik_grad1} and \eqref{eqn:Tik_grad2} together, we have
\begin{align*}
\left[\nabla_{{\cal G}_k} R({\cal A}) \right]_i  &= {\bf a}_k^{(\alpha_{k-1})} {\cal G}_k(j_k) {\bf b}_k^{(\alpha_k)} \\
&= \left( ({\bf b}_k^{(\alpha_k)})^\top \otimes ({\bf e}^{(j_k)})^\top \otimes {\bf a}_k^{(\alpha_{k-1})} \right) \textrm{vec} ({\cal G}_k) ,
\end{align*}
where ${\bf e}^{(j)} \in \mathbb{R}^{n_k}$ denotes the unit vector with the $j$th element equal to $1$ and $0$ otherwise. If we define the $r_{k-1}n_k r_k \times r_{k-1}n_k r_k$ matrix 
\begin{align}\label{eqn:Tik_matrix}
D_k= \left[ \begin{array}{c}
({\bf b}_k^{(1)})^\top \otimes ({\bf e}^{(1)})^\top \otimes {\bf a}_k^{(1)}  \\
({\bf b}_k^{(1)})^\top \otimes ({\bf e}^{(1)})^\top \otimes {\bf a}_k^{(2)} \\
\vdots \\
({\bf b}_k^{(r_k)})^\top \otimes ({\bf e}^{n_k})^\top \otimes {\bf a}_k^{(r_{k-1})}
\end{array}
\right] ,
\end{align}
it follows immediately that $\nabla_{{\cal G}_k} R({\cal A}) = D_k \textrm{vec} ({\cal G}_k)$.

\balance

\section*{Acknowledgment}
Zhongming Chen acknowledges the support of the National Natural Science Foundation of China (Grant No. 11701132). Johan Suykens acknowledges support of ERC AdG A-DATADRIVE-B (290923), KUL: CoE PFV/10/002 (OPTEC); FWO: G.0377.12, G.088114N, G0A4917N; IUAP
P7/19 DYSCO. Ngai Wong acknowledges the support of the Hong Kong Research Grants Council under the General Research Fund (GRF) project 17246416.


\ifCLASSOPTIONcaptionsoff
  \newpage
\fi




\bibliographystyle{IEEEtran}
\bibliography{references}
\end{document}